\documentclass[letterpaper, conference]{ieeeconf}  %

\IEEEoverridecommandlockouts
\usepackage[usenames,dvipsnames,svgnames]{xcolor}
\usepackage{lmodern}
\usepackage{multicol}
\usepackage{cite}
\usepackage{amsmath,amssymb,amsfonts}
\usepackage{algorithmic}
\usepackage{graphicx}
\usepackage{svg}
\usepackage{siunitx}
\usepackage{textcomp}
\usepackage{hyperref}
\usepackage{pifont}
\usepackage{subcaption} 
\usepackage{lipsum}
\usepackage{array}
\renewcommand{\baselinestretch}{0.95}
\usepackage{siunitx}
\usepackage{svg}
\usepackage{float}
\usepackage{dblfloatfix}
\usepackage{textcomp}
\usepackage[linesnumbered,boxed,ruled,vlined]{algorithm2e}
\usepackage{tikz}
\newcommand{\behcetFull}{Beh\c{c}et~A\c{c}\i kme\c{s}e}

\newcolumntype{x}[1]{>{\centering\arraybackslash\hspace{0pt}}p{#1}}

\def\BibTeX{{\rm B\kern-.05em{\sc i\kern-.025em b}\kern-.08em
    T\kern-.1667em\lower.7ex\hbox{E}\kern-.125emX}}

\newcommand{\minimize}{\operatorname{minimize}}

\newcommand{\cfun}[1]{\small\text{#1}}

\newcommand{\norm}[1]{\lVert#1\rVert_2}
\newcommand{\halss}{{\small\textsc{HALSS}}}
\newcommand{\addto}{{\small\textsc{Adaptive{\text -}DDTO}}}
\newcommand{\ddto}{{\small\textsc{DDTO}}}
\newcommand{\optproblem}{{\small\textsc{OptProblem}}}
\newcommand{\subproblem}{{\small\textsc{SubProblem}}}
\newcommand{\baseproblem}{{\small\textsc{BaseProblem}}}
\newcommand{\lcvxproblem}{{\small\textsc{LCvxProblem}}}
\newcommand{\problem}{{\small\textsc{Problem}}}

\newcommand\copyrighttext{
  \footnotesize \textcopyright 2023 IEEE.  Personal use of this material is permitted.  Permission from IEEE must be obtained for all other uses, in any current or future media, including reprinting/republishing this material for advertising or promotional purposes, creating new collective works, for resale or redistribution to servers or lists, or reuse of any copyrighted component of this work in other works.}
\newcommand\copyrightnotice{
\begin{tikzpicture}[remember picture,overlay]
\node[anchor=south,yshift=10pt] at (current page.south) {\fbox{\parbox{\dimexpr\textwidth-\fboxsep-\fboxrule\relax}{\copyrighttext}}};
\end{tikzpicture}}
\title{HALO: Hazard-Aware Landing Optimization for Autonomous Systems}

\author{Christopher R. Hayner$^{\star}$, Samuel C. Buckner$^{\star,\dag}$, Daniel Broyles$^{\ddag}$,\\ Evelyn Madewell, Karen Leung, \behcetFull
\thanks{$^\star$Both authors contributed equally.}
\thanks{$\dag$ Any opinion, findings, and conclusions or recommendations expressed in this material are those of the authors(s) and do not necessarily reflect the views of the National Science Foundation.}
\thanks{$\ddag$ The views expressed in this article are those of the author and do not reflect the official policy or position of the United States Air Force, Department of Defense, or the U.S. Government.}
\thanks{The authors would like to acknowledge Annika Singh and Michael Hayner.}
\thanks{Dept. of Aeronautics and Astronautics, University of Washington, USA. Emails: {\tt\small \{haynec, sbuckne1\}@uw.edu}}%
} 
\graphicspath{{./figures/}}

\begin{document}

\maketitle
\copyrightnotice
\begin{abstract} 
With autonomous aerial vehicles enacting safety-critical missions, such as the Mars Science Laboratory Curiosity rover’s landing on Mars, the tasks of automatically identifying and reasoning about potentially hazardous landing sites is paramount. 
This paper presents a coupled perception-planning solution which addresses the hazard detection, optimal landing trajectory generation, and contingency planning challenges encountered when landing in uncertain environments. 
Specifically, we develop and combine two novel algorithms, Hazard-Aware Landing Site Selection (HALSS) and Adaptive Deferred-Decision Trajectory Optimization (Adaptive-DDTO), to address the perception and planning challenges, respectively.
The HALSS framework processes point cloud information to identify feasible safe landing zones, while Adaptive-DDTO is a multi-target contingency planner that adaptively replans as new perception information is received.
We demonstrate the efficacy of our approach using a simulated Martian environment and show that our coupled perception-planning method achieves greater landing success whilst being more fuel efficient compared to a non-adaptive DDTO approach.

\end{abstract}
\section{Introduction}
\label{sec:introduction}
Autonomous robotic systems are poised to carry out a number of safety-critical missions, such as planetary exploration and disaster response, that otherwise would be dangerous, expensive, or impractical for humans to perform. In the case of autonomous aerial robots (e.g., unoccupied aerial vehicles), where landing safely is critical to mission success, the robots must reason about terrain conditions to identify and avoid potentially hazardous landing areas which threaten the mission goals.

In this work, we propose a perception-aware trajectory planning algorithm that enables an autonomous aerial robot to land safely in unknown or uncertain environments. Specifically, our algorithm combines a LiDAR-based hazard-aware landing site selector with an adaptive multi-target trajectory optimizer to generate fuel-efficient trajectories that are robust to changes in landing sites priority order as new observations are received.

\begin{figure}[!t]
    \includegraphics[width=0.48\textwidth]{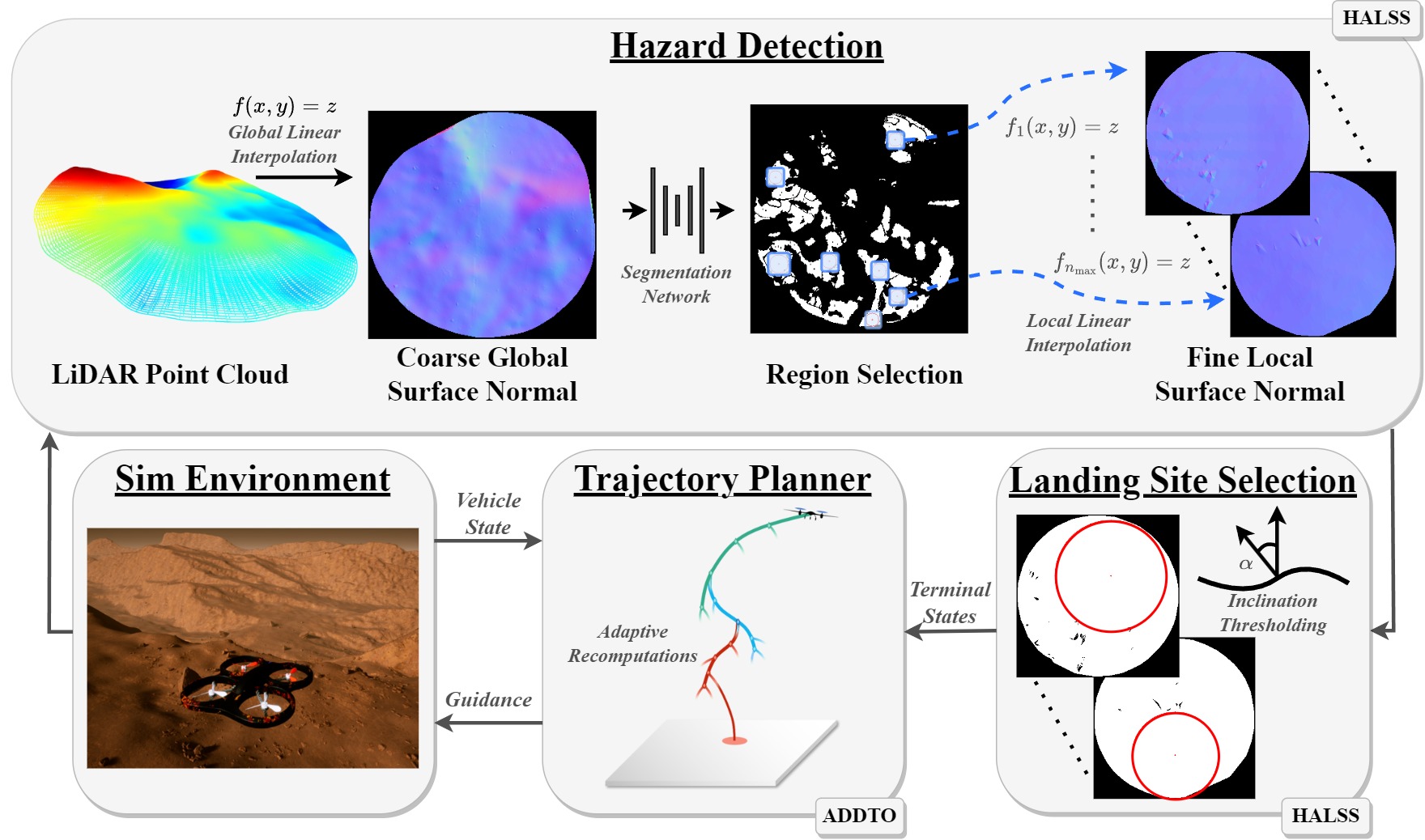}
    \caption{System Overview: \halss \ identifies viable landing sites used in the multi-target trajectory optimization generated by \addto. The resulting guidance is used to command a (simulated) quadrotor, and landing site viability is continually evaluated as new measurements are processed.}
    \label{fig:sys_overview}
\end{figure}

\noindent \textbf{Related work.} Hazard detection is a critical element of NASA's efforts to develop precision landing technologies for planetary exploration missions \cite{Epp2008-hj}. The NASA Safe and Precise Landing - Integrated Capabilities Evolution (SPLICE) project specifically outlines next-generation hazard detection capabilities, which primarily operate on lidar-derived digital elevation models (DEMs) to measure the slope and roughness of potential landing sites \cite{Restrepo2020-do}. These DEM-based geometric approaches to hazard classification include probabilistic modeling \cite{ivanov2013,brockers2021hd, brokers2022landing} and fuzzy-reasoning-based safety maps \cite{Konstantinidis2020-jg}. However recent work has achieved performance improvements over classical methods via the use of deep learning, including landing site hazard identification from semantic segmentation networks \cite{Moghe2020-fq,Tomita2020-qa} and Bayesian deep-learned segmentation and prediction variance for uncertainty-aware safe landing site identification \cite{Tomita2021}. Our proposed method leverages the benefits of Bayesian segmentation with the theoretical guarantees of a direct geometric representation of hazardous terrain. 

An important consideration in perception-aware trajectory planning is the management of contingency plans when a previously-viable target state is found to be infeasible or impractical due to map and localization uncertainty, fine-grain objects not initially detected by LiDAR or other on-board sensors, and uncertainty in the environment. Model predictive control (MPC) approaches have demonstrated robust performance for trajectory guidance in dynamic and complex landing scenarios  \cite{greer2020shrinking}, however MPC does not consider multiple contingency options. Deferred-decision trajectory optimization (\ddto) \cite{elango2022deferring} offers a framework to compute trajectories to multiple selected targets while maximizing the deferral time before needing to commit to a target for a given suboptimality tolerance. \ddto\space is notably distinguished from stochastic methods to handle uncertainty in that it is entirely deterministic and requires no probabilistic knowledge of the environment. The trajectory generation method presented in this paper (\addto) combines the advantages of both DDTO and MPC. \\
\noindent \textbf{Contributions.} 
Contributions of this work include novel frameworks for determining safe landing sites in uncertain and hazardous areas, known as Hazard-Aware Landing Site Selection (\halss), and for adaptive multi-target trajectory optimization, referred to as (\addto).\\
\noindent \textbf{Organization.} 
Sections \ref{architecture_section} and \ref{sim_env_section} discuss the proposed system architecture and simulation environment used for this research, respectively. The \halss \: framework is described in Section \ref{sec:hazard_detection}, and Section \ref{sec:trajoptimization} covers the \addto \: algorithm. Results from an integrated simulation and comparative study are presented in Section \ref{sec:results}, followed by conclusions in Section \ref{sec:conclusion}. 
\section{Architecture} 
\label{architecture_section}
Fig. \ref{fig:sys_overview} depicts our proposed solution at a high level. The \halss\ algorithm performs the hazard detection from LiDAR measurements of the terrain and determines safe landing sites represented by circular regions. These landing sites are passed as terminal states to \addto, which removes landing sites as they are found to be unsafe or are no longer reachable as the vehicle passes by its respective branch point in the trajectory. \addto\ then solves the landing optimization problem, and the resulting multi-target trajectory is executed by a low-level controller. The landing sites are returned to \halss\ to have their radii updated if any new unsafe terrain are identified within. This loop is repeated until the autonomous system reaches a cutoff altitude, at which point it locks to the best landing site.
\section{Hazard-Aware Landing Site Selection}
\label{sec:hazard_detection}

In this section, we introduce the \halss\:framework that selects safe landing sites from point cloud data.
Notably, \halss\ performs hazard detection at multiple scales which allows for efficient computation without sacrificing hazard detection performance. We define the global scale as encompassing the entirety of the observed map and local scale as smaller subsets of the observed map. \halss\ first identifies larger-scale hazards by performing coarse hazard detection on the global scale through the process outlined in \ref{sec:coarse_hd}. From the resulting global safety map, it selects regions to identify potential landing sites using the process outlined in \ref{sec:landing_site_selection}. For each region identified by the coarse hazard detection, fine hazard detection is performed to identify smaller-scale hazards using the process outlined in \ref{sec:fine_hd}. From each of the identified local safety maps, a landing site is selected using the process outlined in \ref{sec:landing_site_selection}.

\subsection{Coarse Hazard Detection}
\begin{figure}
    \includegraphics[width=0.48\textwidth]{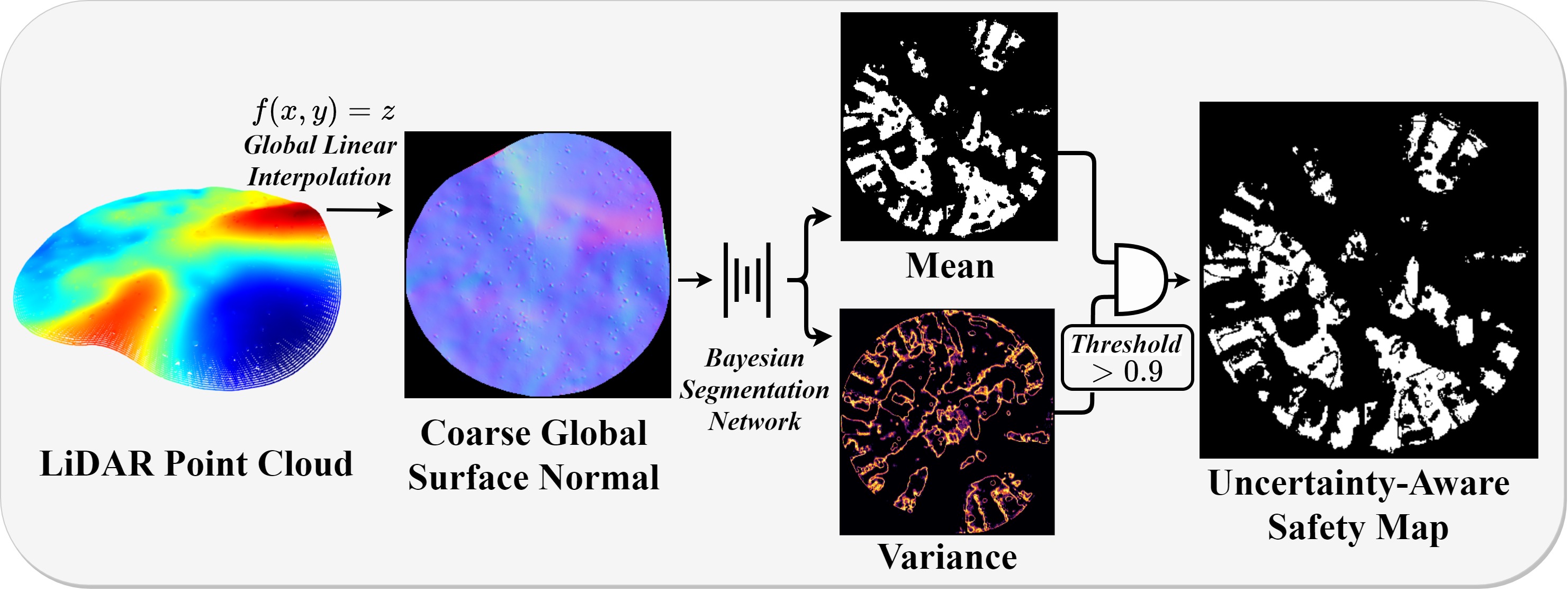}
    \caption{Coarse Hazard Detection Process: a LiDAR-derived surface normal image is fed into a Bayesian SegNet which classifies flat and sloped/rocky regions; the resulting output is an uncertainty-aware safety map with white pixels corresponding to safe/flat areas.}
    \label{fig:coarse_hd}
\end{figure}

Coarse hazard detection aims to identify landing hazards, such as large slopes and rocks, in order to generate a global safety map. This task is accomplished via a Bayesian segmentation network for classifying terrain as safe/unsafe. This section discusses the coarse hazard detection pipeline, as shown in Fig. \ref{fig:coarse_hd}.
\label{sec:coarse_hd}
\subsubsection*{Global Surface Normal Interpolation} \label{sec:global_sn_interp} The surface normal representation of the terrain is used as the input to the segmentation network due to its ability to express terrain slope and roughness features. It consists of three-channel data which encodes the unit surface normal vector for each pixel and can be derived from LiDAR point cloud data (as shown in Fig. \ref{fig:coarse_hd}). To create a global surface normal image from a point cloud, we first cull points to decrease the interpolation time using a roughness-preserving downsampling algorithm. This algorithm constructs a two-dimensional uniform grid where, for each cell, the points with the largest and the smallest $Z$-value are chosen, and the remaining points are culled. Choosing the largest and smallest Z-value points ensures that the roughness is preserved in the culled point cloud, which is the most conservative case and results in the least safe terrain. A function is then linearly interpolated from the culled point cloud and evaluated at uniform grid spacing.
\subsubsection*{Network Architecture}
We use a Bayesian segmentation network based on the Bayesian Seg-Net  architecture \cite{Kendall2015bayeseg}, which is well suited to learn higher dimensional features (such as larger slopes and rocks). Monte Carlo (MC) dropout layers are included after each series of convolution operations to account for uncertainty in the predictions. To produce a binary safety map, we apply a sigmoid transform to the output of the network. During inference, the global surface normal is propagated through the network, resulting in 30 stochastic samples. The mean and variance of these outputs model the predicted safety and approximated Bayesian uncertainty of that prediction, respectively, similar to the process outlined in \cite{Tomita2021}. We then normalize and threshold the variance at a conservative value of 0.9 which was experimentally chosen and apply the logical AND operation to the variance and mean, resulting in the uncertainty-aware safety map.

\subsection{Fine Hazard Detection}
\label{sec:fine_hd}
\begin{figure}
\centering
    \includegraphics[width=0.45\textwidth]{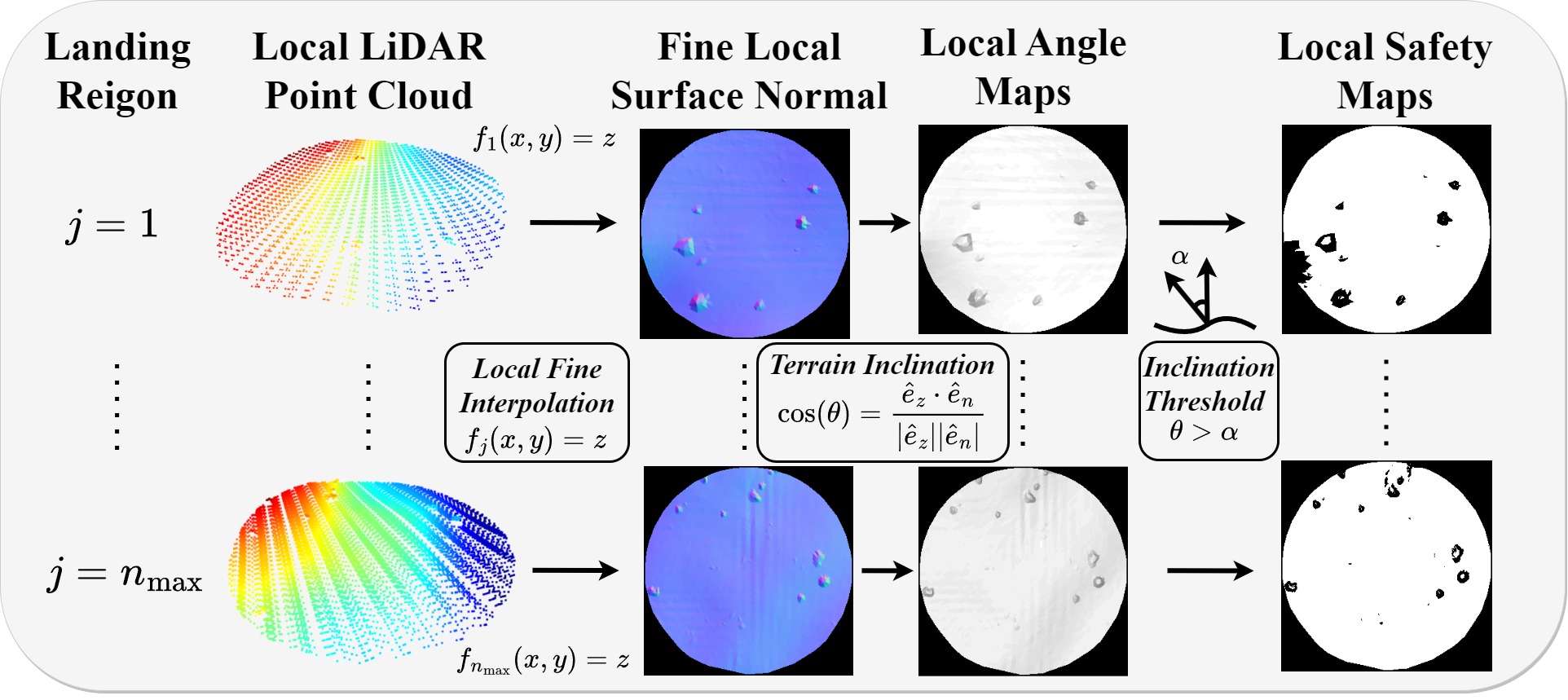}
    \caption{Fine Hazard Detection Process: local safety maps are obtained via angle thresholding high-resolution local surface normals for each landing region.}
    \label{fig:fine_hd}
\end{figure}

Potential landing sites identified from the global safety map are further analyzed to identify small hazards and create detailed, high-resolution safety maps for each landing region. This section explains the fine hazard detection process, as depicted in Fig. \ref{fig:fine_hd}.

\subsubsection*{Localized Surface Normal Interpolation}
In order to ensure that smaller localized details are captured with respect to each region, a subset of the original un-culled point cloud within the radius (plus a buffer) of the center coordinate of the identified regions is taken, resulting in a set of localized point clouds, as seen in Fig. \ref{fig:fine_hd}. To prevent the computation time from growing unbounded as new points are added to the global point cloud, each of these point clouds is then independently downsampled using the same roughness-preserving downsampling technique introduced in \ref{sec:global_sn_interp}, but with finer grid spacing. Similar to the global surface normal, the localized surface normals are created by linearly interpolating a function to each culled local point cloud, sampling at even grid spacing, and finally calculating the surface normals.

\subsubsection*{Inclination Thresholding} 
To identify smaller and more low-dimensional hazards, a pure geometric representation of the slope of the terrain is well suited as it offers explainable representations of smaller hazards. The local angle map is obtained by taking the dot product between the unit surface normal vector, $\hat{e}_n\in \mathbb{R}^3$, from each local point cloud with the vertical unit vector, $\hat{e}_z\in \mathbb{R}^3$, at each point in the local surface normal map. The local angle map, $\theta\in \mathbb{R}$, is then thresholded against the maximum allowable inclination, $\alpha\in \mathbb{R}$, such that $\theta > \alpha$ implies an unsafe region, which results in the local safety maps.

\subsection{Region and Landing Site Selection} \label{sec:landing_site_selection}
\begin{figure}[h]
    \vspace{-2mm}
    \includegraphics[width=0.48\textwidth]{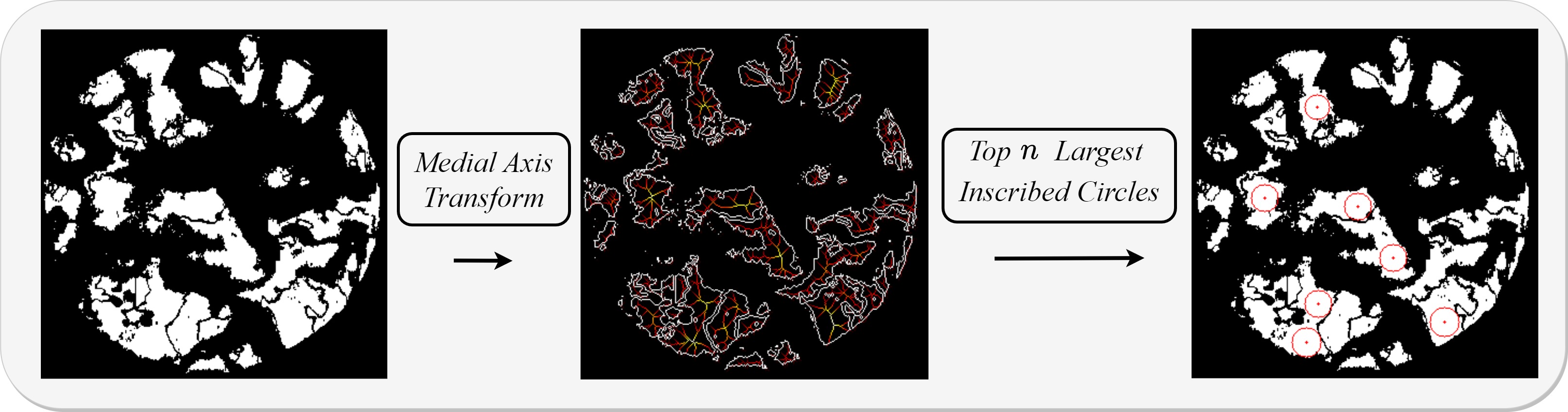}
    \caption{Example of safe region (red circles) identification via medial axis skeletonization of the binary safety map.  }
    \label{fig:lss}
\end{figure}

Potential safe regions and landing sites are identified by considering the largest inscribed circle contained in the safe region, which is easily obtained from the medial axis transform (MAT) \cite{Choi1997-lo}. The medial axis is a continuous version of the Voronoi diagram, and in the 2-D domain can be thought of as the set of centers of maximally inscribed discs with respect to some boundary. Applying the MAT to a binary image results in a skeleton, which is an array of values representing the distance to a boundary in the original object; the boundary in this particular case is the set of black pixels. Fig. \ref{fig:lss} shows an example of medial axis skeletonization of a safety map. To find a set of candidate landing sites, the global safety map skeleton is used to quickly and efficiently find the top $n$ (number of desired landing targets) largest circular safe zones. For fine hazard detection, this process is repeated for each local safety map with $n=1$.

\subsubsection*{Scoring Parameters}\label{scoring}
Once obtained, each landing site is scored using the following metrics: the point cloud density within the landing site (\textit{density}), the Euclidean distance between the landing site and vehicle in the NED frame (\textit{proximity}), the proximity to other landing sites (\textit{clustering}), the $99^{th}$ percentile of the uncertainty map within landing site (\textit{uncertainty}), and the landing site bounding circle radius (\textit{size}).
\section{Adaptive Deferred-Decision Trajectory Optimization}
\label{sec:trajoptimization}

In the context of hazard-awareness, acquired targets can lose viability at a future time due to navigation errors and environmental uncertainty, thus reducing the set of available targets. In this section, a novel algorithm is introduced to maintain a bounded number of targets in the reachable set of the vehicle for path planning at all times. This approach starts with (1) the baseline trajectory optimization problem formulation, and builds on it with (2) a single-shot multi-targeted path planning approach (\ddto), which is enhanced to an adaptively-recomputed multi-targeted path planning approach (\addto).

\subsection{Baseline Problem Formulation}
\label{subsec:baseline_prob}

To model the vehicle dynamics (in the context of a multirotor), a three degree-of-freedom (3-DOF) state-space model is considered of the form:
\vspace*{-1mm}
\begin{equation} \label{eq:ss}
    \dot{x}(t) = Ax(t) + Bu(t) + p
\end{equation} where $r(t) \in \mathbb{R}^3$ is the vehicle's position, $\dot{r}(t) \in \mathbb{R}^3$ is the vehicle's velocity, and $x(t) = \left[r(t)^\top, \dot{r}(t)^\top\right]^\top\in\mathbb{R}^{6}$ is the vehicle's state. Furthermore, the control signal $u(t) \in \mathbb{R}^3$ is the commanded thrust of the vehicle (henceforth denoted as $T(t)$). Consequently, the vehicle's simplified dynamics can be described as a double integrator system with an affine gravity term, such that:
\begin{equation} \label{eq:ss_matrices}
    A = 
    \left[\begin{array}{cc}
        0 & I \\
        0 & 0
    \end{array}\right], \
    B = 
    \left[\begin{array}{c}
        0  \\
        \frac{I}{m}
    \end{array}\right],\
    p = 
    \left[\begin{array}{c}
        0  \\
        -g
    \end{array}\right]
\end{equation}where $g \in \mathbb{R}^3$ is the acceleration due to gravity, $m \in \mathbb{R}$ is the mass of the vehicle, and $I\in\mathbb{R}^{3\times 3}$ is the identity matrix. We consider a surface-fixed inertial coordinate frame consisting of basis vectors $e_j \in \mathbb{R}^3$, for $j=1,2,3$, which correspond to downrange, crossrange and altitude, respectively. The state-space model \eqref{eq:ss_matrices} assumes that gravity is constant and neglects aerodynamic drag.

The objective of the single-shot trajectory optimization problem (\baseproblem) is to guide the target from its current airborne state, $x_0$, to a safe target zone on the ground, $x_f$, subject to the dynamics in \eqref{eq:ss}, as well as constraints on the state $x(t)$ and control $u(t)$, while minimizing control effort (cumulative thrust):

\vspace{2mm}
\noindent\fbox{%
\parbox{0.98\linewidth}{%

\begin{equation}
    \baseproblem \text{ (3-DOF Non-Convex)} \tag{\ref{eq:baseline_prob}}
\end{equation}
\vspace{-5mm}
\begin{subequations}\label{eq:baseline_prob}
\begin{alignat}{2}
    &\underset{T(\cdot)}{\minimize} &\qquad& \int_0^{t_f} \norm{T(\tau)} d\tau \label{eq:baseline_cost}\\
    &\text{subject to}     &      & \dot{x}(t) = Ax(t) + Bu(t) + p, \label{eq:baseline_c1}\\
    &                      &      & 0 < T_{\mathrm{min}} \leq \norm{T(t)} \leq T_{\mathrm{max}}, \label{eq:baseline_c2}\\
    &                      &      & \norm{T(t)}\cos(\gamma_{\mathrm{max}}) \leq e_3^\top T(t), \label{eq:baseline_c3}\\
    &                      &      & \norm{\dot{r}_{1:2}(t)} \leq v_{\mathrm{max}}^{\mathrm{lat}}, \label{eq:baseline_c4}\\
    &                      &      & \text{abs}({\dot{r}_3(t)}) \leq v_{\mathrm{max}}^{\mathrm{vert}}, \label{eq:baseline_c5}\\
    &                      &      & x(0) = x_0, x(t_f) = x_f \label{eq:baseline_c6}
\end{alignat}
\end{subequations} 
}
\vspace{-2mm}
}
\vspace{1mm}

In problem \eqref{eq:baseline_prob}, cumulative commanded thrust is minimized for in the cost function \eqref{eq:baseline_cost}, with constraints for dynamics \eqref{eq:baseline_c1}, lower ($T_{\mathrm{min}}$) and upper ($T_{\mathrm{max}}$) bound constraints on thrust magnitude \eqref{eq:baseline_c2}, a tilt angle bound $\gamma_{\mathrm{max}}$ to the vertical $e_3$ \eqref{eq:baseline_c3}, a lateral velocity bound $v_{\mathrm{max}}^{\mathrm{lat}}$ \eqref{eq:baseline_c4}, a vertical velocity bound $v_{\mathrm{max}}^{\mathrm{vert}}$ \eqref{eq:baseline_c5}, and initial and terminal boundary conditions \eqref{eq:baseline_c6}, across a fixed time-of-flight denoted $t_f\in\mathbb{R}$.

The lower bound in \eqref{eq:baseline_c2} causes the constraint, and therefore the problem, to become non-convex. The lossless convexification (LCvx) approach (see \cite{accikmecse2011lossless,malyuta2021convex}) provides the capability to convexify these constraints in a way that is both lossless (preserves global optimality) and provides feasibility guarantees. This can be done by introducing the slack variable $\Gamma(t) \in \mathbb{R}$, which substitutes $\norm{T(t)}$ in the cost \eqref{eq:baseline_cost} and constraints \eqref{eq:baseline_c2}--\eqref{eq:baseline_c3} and provides an additional relaxation constraint $\norm{T(t)} \leq \Gamma(t)$. This augmentation to \baseproblem\ will be used throughout the remainder of this paper and referred to as \lcvxproblem. The 3-DOF solution that \lcvxproblem\ provides can be considered a proxy to a full 6-DOF solution (with yaw unconstrained) due to the differential flatness property of multirotor vehicles \cite{fliess1995flatness, nguyen2017reliable}.

\subsection{Adaptive Deferred-Decision Trajectory Optimization} 
\label{subsec:DDTO}

\begin{figure}[!t]
    \centering
    \includegraphics[width=\linewidth]{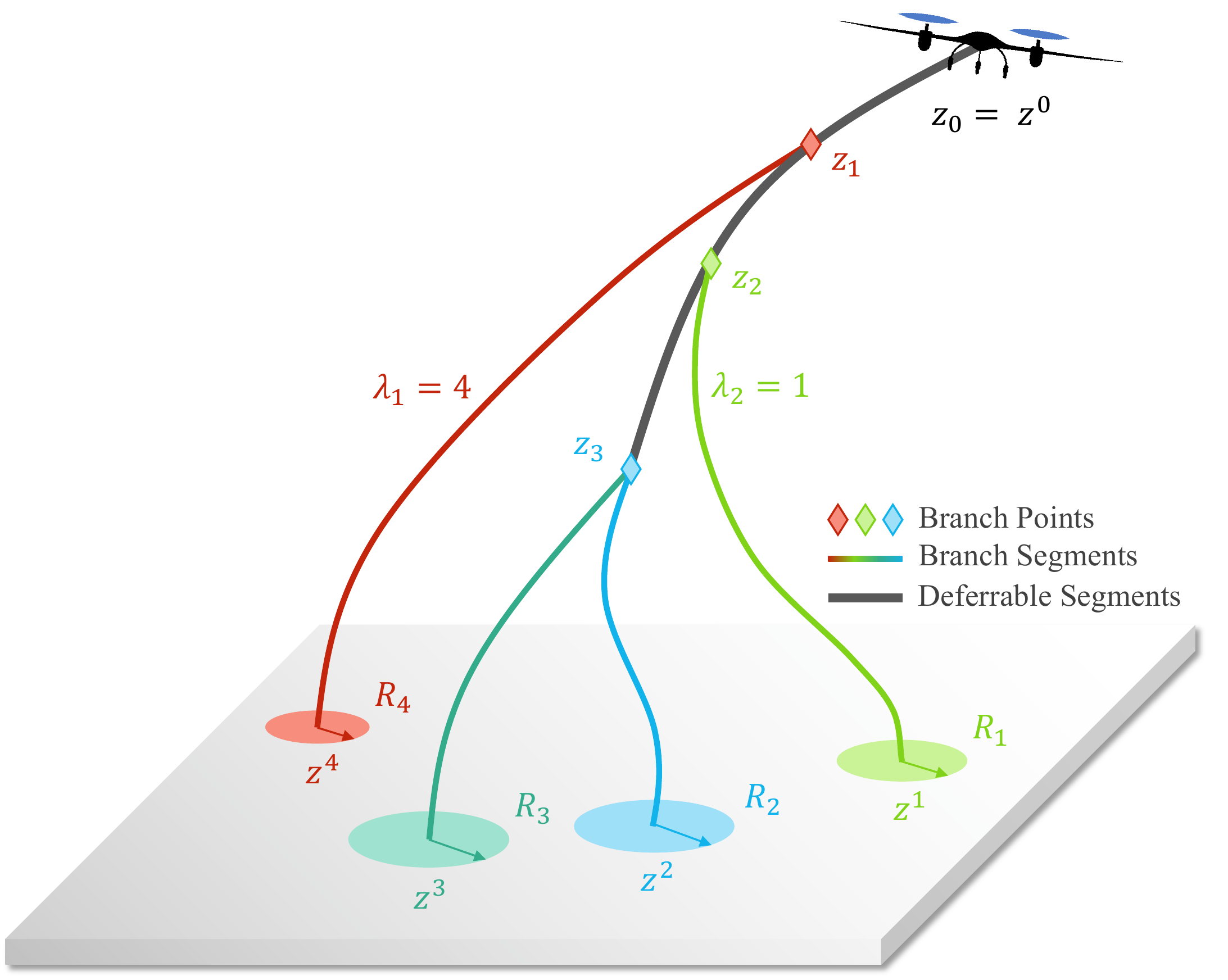}
    \caption{Illustration of a \ddto\space solution to the multirotor landing problem with $n = 4$ targets.}
    \label{fig:DDTO_Illustration}
\end{figure}

\begin{figure}[!t]
    \centering
    \includegraphics[width=\linewidth]{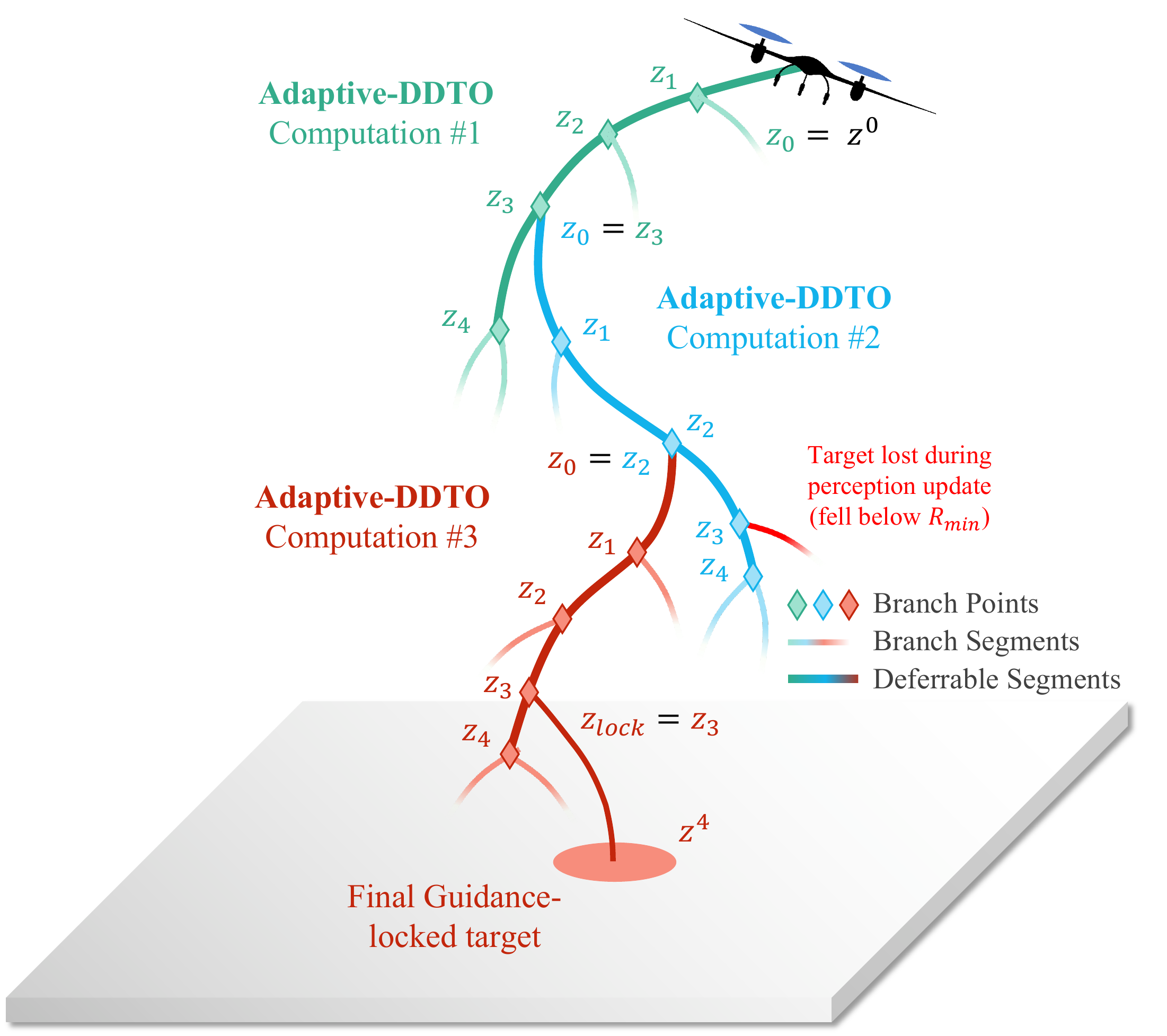}
    \caption{Illustration of an \addto\space solution to the multirotor landing problem with $n_{\mathrm{min}} = 3$ and $n_{\mathrm{max}} = 5$.}
    \label{fig:ADDTO_Illustration}
\end{figure}

Deferred-decision trajectory optimization (\ddto) \cite{elango2022deferring} offers an enhancement to \lcvxproblem\space to enable multi-target path planning while keeping multiple target states in the reachable set of the vehicle and maximizing the deferral time to each of them given a set of suboptimality tolerances (relative to the optimal single-shot \lcvxproblem\space solution to the same target) and a target rejection sequence (see Fig. \ref{fig:DDTO_Illustration}). The \addto\space algorithm builds on this idea to provide a fault-tolerant trajectory optimization framework to enable the vehicle to maintain at least $n_{\mathrm{min}} \in \mathbb{N}$ targets and up to $n_{\mathrm{max}} \in \mathbb{N}$ targets in consideration at all times (see Fig. \ref{fig:ADDTO_Illustration} and Algo. \ref{alg:addto}). The following design can be considered similar to a shrinking-horizon MPC problem (e.g., \cite{greer2020shrinking}) which is conditionally executed after non-uniform time intervals. The following paragraphs will describe the prerequisite knowledge behind Algorithm \ref{alg:addto}.\footnote{Note that in Algorithm \ref{alg:addto}, when calling the underlying \ddto\space algorithm described in \cite{elango2022deferring}, \lcvxproblem\space is used as the \optproblem, with the corresponding constraint set used to solve each \subproblem.}

Consider a state-space system with dimensions $n_x,n_u \in \mathbb{N}$ corresponding to the number of states and control inputs, respectively. The vehicle's initial state is denoted $z^0 \in \mathbb{R}^{n_x}$, with $n \in \mathbb{N}$ acquired target states $z^j \in \mathbb{R}^{n_x}$ for $j \in J$, where $J \subset \mathbb{N}$ is a finite indexing set for identifying the $n$ targets. Additionally, there exist $n-1$ branch points $\{z_k\}_{k=1}^{n-1} \subset \mathbb{R}^{n_x}$ with associated time horizons $\{\tau\}_{k=1}^{n-1} \subset \mathbb{R}$, from which $n-1$ deferrable segments exist between each $z_{k-1}$ and $z_k$, starting with $z_0 = z^0$, along with suboptimality tolerances $\{\epsilon\}_{k=1}^{n}$ for each target. Each target also has a bounding circle associated with the safety map (see Section \ref{sec:hazard_detection}) that is updated at each branch point with a radius $R^j \in \mathbb{R}$. These radii will be implicitly considered a (dynamically-updated) parameter of $z^j$ to simplify notation going forward (along with the scoring parameters discussed in Section \ref{sec:landing_site_selection}).

\ddto\space also requires the specification of a target rejection order $\{\lambda_k\}_{k=1}^{n-2} \subset \mathbb{N}$, where identification tags from $J$ are specified in the order of deferral. To facilitate this, a desirability score is assigned to each target and represented by the evaluation function $g_{\mathrm{des}} : \mathbb{R}^{n_x} \rightarrow [0,\infty)$. A proposed heuristic for this evaluation function is a weighted sum $g_{\mathrm{des}}(z^j) = w_{\mathrm{des}}^\top c_{\mathrm{des}}$, where $c_{\mathrm{des}} \in \mathbb{R}^5$ contains the scores related to point cloud density, proximity to vehicle, site clustering, network prediction uncertainty, and landing zone size, as described in Section \ref{sec:landing_site_selection}, and $w_{\mathrm{des}} \in \mathbb{R}^5$ contains corresponding tunable weights for each element of $c_{\mathrm{des}}$.

The function $g_{\mathrm{rem}} : \mathbb{R}^{n_x} \rightarrow \{0,1\}$ determines if a target should be removed or not, and returns \texttt{true} if the target bounding radii exceeds the minimum safe bounding radius, denoted $g_{\mathrm{rem}}(z^j) = R^j < R_{\mathrm{min}}$.

The switching decision at a branch point is represented by a function $g_{\mathrm{switch}} : \mathbb{R}^{n_x} \rightarrow \{0,1\}$ and returns \texttt{true} if the desirability score associated with the target in consideration for deferral exceeds all other targets' scores, such that $g_{\textrm{switch}}(z^j) = g_{\textrm{des}}(z^j) > \cfun{\textrm{max}}(\{g_{\textrm{des}}(z^i)\}_{i \in \{J/j\}})$

Under certain conditions, it may be desirable to \textit{lock} the guidance to the best available target $z^{j_{\mathrm{lock}}}$, disabling \addto\space re-computations and applying control towards this target at the current state $z_{\mathrm{lock}}$ (updated implicitly) with current time $\tau_{\mathrm{lock}}$ until the terminal condition $z^{j_{\mathrm{lock}}}$ is reached. A reasonable locking condition based on a minimum ``cutoff" altitude $h_{\mathrm{cutoff}}$ could be used, such that $g_{\mathrm{lock}} : \mathbb{R} \rightarrow \{0,1\}$ is defined with $g_{\mathrm{lock}}(h) = h < h_{\mathrm{cutoff}}$. To index into the altitude of $z_k$, the vector $e_h = [0, 0, 1, 0 , 0, 0]^\top$ is used.

There are also several other useful functions and relations that will be considered for the \addto\space formulation:
\begin{itemize}
    \item $h_{\mathrm{acquire}} : \mathbb{N} \rightarrow \mathbb{R}^{n_x}$ calls from \halss\space to acquire $n_{\mathrm{max}} - n$ additional targets, where $n \geq n_{\mathrm{min}}$ is the current number of targets.
    \item $h_{\mathrm{update}} : \mathbb{R}^{n_x} \rightarrow \mathbb{R}^{n_x}$ calls from \halss\space to update target $z^j$ and associated properties.
    \item $\cfun{bisect} :~$\problem$~\rightarrow \mathbb{N}$ performs a minimum-feasibility bisection search across the time-of-flight $t_f\in\mathbb{R}$ for a fixed time-of-flight problem (such as \lcvxproblem) to return the optimal discretization node count $N^*$ for a given discretization time step $\Delta t\in\mathbb{R}$ (such that $t_f^* = N^* \Delta t$).
    \item $\cfun{sort\_perm}_n : \mathbb{R} \rightarrow \mathbb{N}$ sorts the input collection of cardinality $n \in N$ in monotonically increasing order and returns the permutation indices that indicate the sorting order.
    \item $\cfun{comm\_control} : (\mathbb{R}^m \times \mathbb{N} \times  \mathbb{R} \times \mathbb{R}) \rightarrow \varnothing$ extracts and commands the control sequence from \ddto-returned concatenated vector $V \in \mathbb{R}^m$ (where $m$ is the total number of control elements, see \cite{elango2022deferring}) towards target $j$ between time $t_1$ and time $t_2$.
\end{itemize}

Algorithm \ref{alg:addto} notably contains recursive calls to itself for \textit{adaptive} re-computations when $n < n_{\mathrm{min}}$.

\begin{algorithm}
\DontPrintSemicolon
\SetArgSty{text}
\SetKwInput{KwData}{Input}
\SetKwInput{KwResult}{Return}
    \KwData{$z^0, J, \{z^j\}_{j \in J}$}
    \tcc{Begin initialization}
    $z_0 \gets z^0$\;
    $\{z^j\}_{j=1}^{n_{\mathrm{max}}} \gets h_{\mathrm{acquire}}(n_{\mathrm{max}} - |J|)$\;
    $n \gets n_{\mathrm{max}}$\;
    $J \gets 1:n$\;
    $\{\lambda_k\}_{k=1}^{n-2} \gets \cfun{sort\_perm}_n(\{g_{\mathrm{des}}(z^j)\}_{j \in J})$\;
    \tcc{End initialization}
    \For{$j \in J$}{
        $N^j \gets \cfun{bisect}(\lcvxproblem)$\;
    }
    $\{\tau_k\}_{k=1}^{n-1}, \{z_k\}_{k=1}^{n-1}, V \gets \ddto(J, \{\lambda_k\}_{k=1}^{n-2}, z_0, \{z^j\}_{j \in J}, \{\epsilon^j\}_{j \in J}, \{N^j\}_{j \in J})$\;
    \For{$k=1,...,n-1$}{
        $z^j \gets h_{\mathrm{update}}(z^j,k)$\;
        \If{$g_{\mathrm{switch}}(z^j)$}{
            $J \gets \{\lambda_k\}$\;
        } \Else {
            $J \gets \{J/\lambda_k\}$\;
        }
        \For{$j \in J$}{
            \If{$g_{\mathrm{rem}}(z^j)$}{
                $J \gets \{J/j\}$\;
            }
        }
        $n \gets |J|$\;
        \If{$n < n_{\mathrm{min}}$}{
            
            $\addto(z_k, J, \{z^j\}_{j \in J})$\;
        }
        \If{$g_{\mathrm{lock}}(e_h^\top z_k)$}{
            $j_{\mathrm{lock}} \gets \cfun{argmax}(\{g_{\mathrm{des}}(z^j)\}_{j \in J})$\;
            $\cfun{comm\_control}(V, j_{\mathrm{lock}}, \tau_{k-1}, N^{j_{\mathrm{lock}}}\Delta t)$\;
        } \Else {
            $\cfun{comm\_control}(V, \lambda_k, \tau_{k-1}, \tau_k)$\;
        }
    }
\caption{\addto}
\label{alg:addto}
\end{algorithm}

\section{Results} \label{sec:results}
This section discusses implementation details and experimental results from an integrated simulation of a combined \halss\space \addto\space implementation, as well as analytical results from a comparative study between \addto\space and \ddto. \vspace*{-2mm}
\begin{figure*}[!t]
    \centering
    \begin{subfigure}{.25\textwidth}
      \centering
      \includegraphics[width=.98\linewidth]{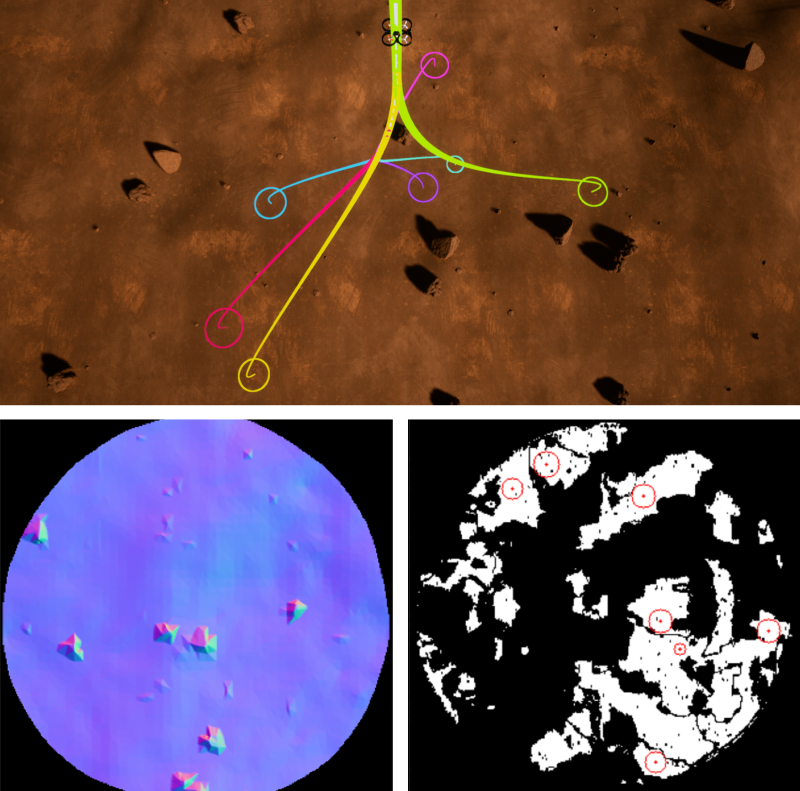}
    \end{subfigure}%
    \begin{subfigure}{.25\textwidth}
      \centering
      \includegraphics[width=.98\linewidth]{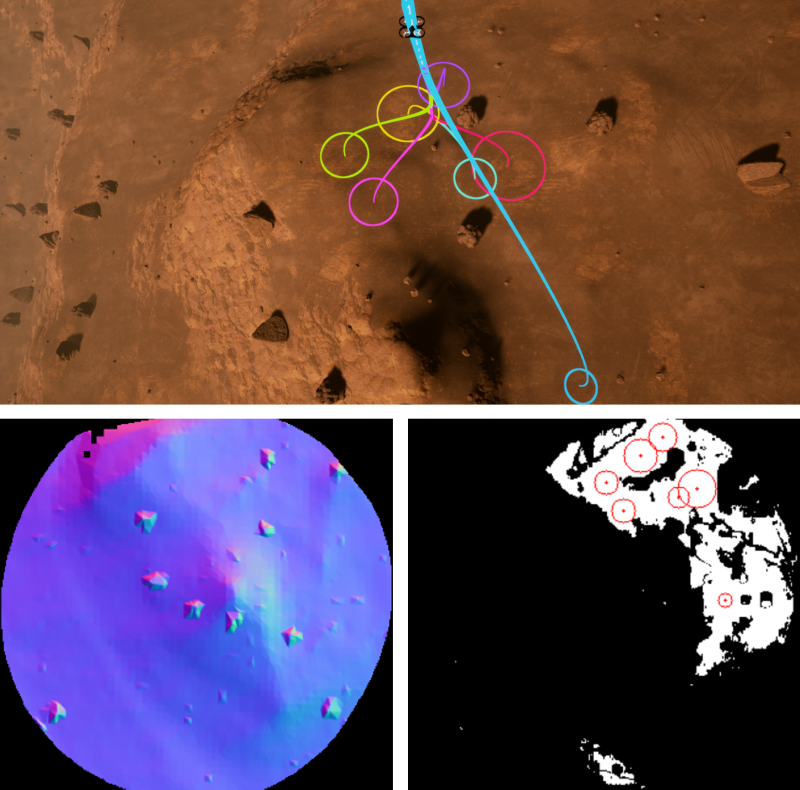}
    \end{subfigure}%
    \begin{subfigure}{.25\textwidth}
      \centering
      \includegraphics[width=.98\linewidth]{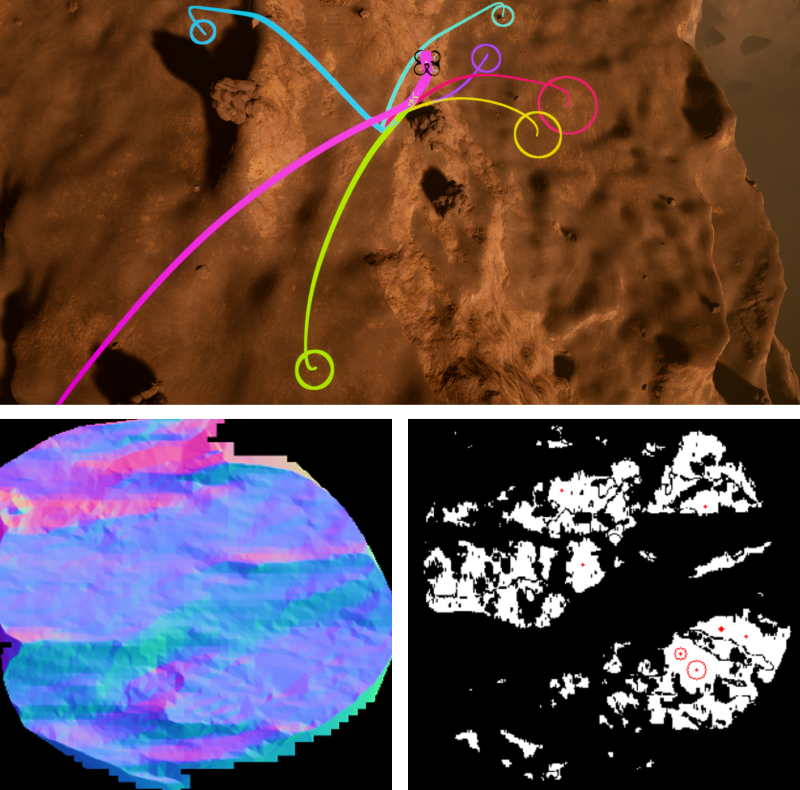}
    \end{subfigure}%
    \begin{subfigure}{.25\textwidth}
      \centering
      \includegraphics[width=.98\linewidth]{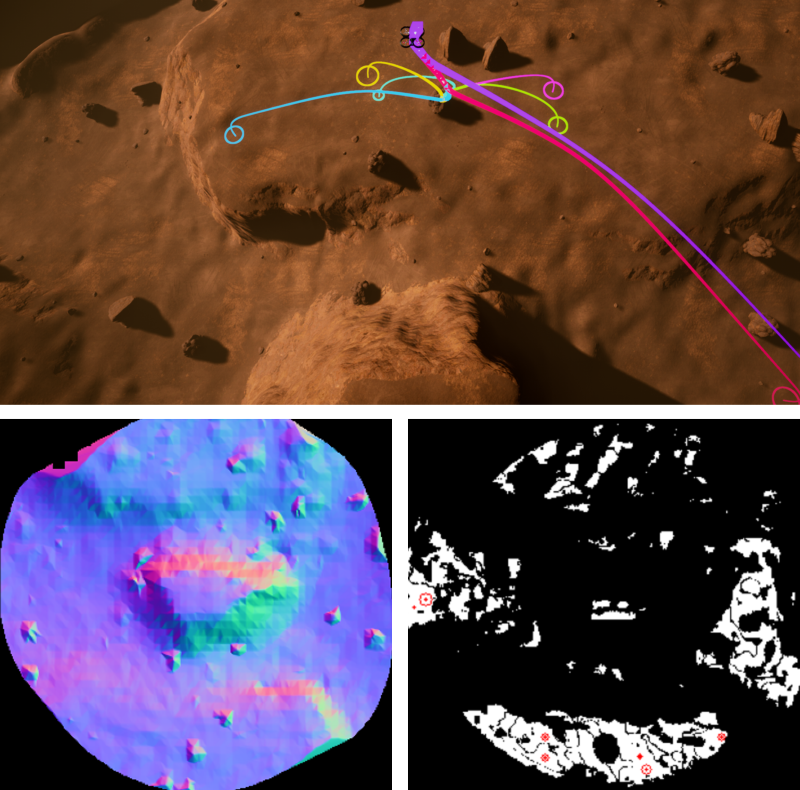}
    \end{subfigure}%
    \caption{Snapshots of four simulation instances on different terrain maps of increasing complexity from left to right. The top images shows our simulation environment during a run with the \ddto\ branches and landing sites, the bottom images show the global surface normal (left) and safety (right) maps with associated landing sites (red circles). }
    \label{fig:airsimdemo_snapshots}
\end{figure*}

\begin{figure*}[!t]
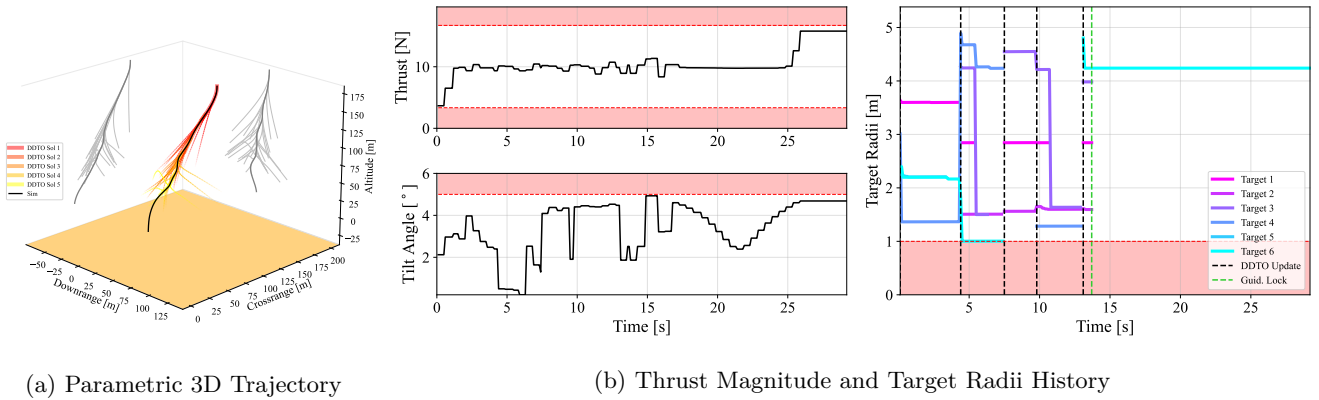

    \vspace{3.5mm}
    \begin{subfigure}{.2875\textwidth}
      \centering
      \includesvg[width=1\linewidth]{airsim_integration/airsimdemo_parametric_3D_trajectory}
      \caption{Parametric 3D Trajectory}
      \label{fig:airsimdemo_parametric_3D_trajectory}
    \end{subfigure}%
    \begin{subfigure}{.7125\textwidth}
      \centering
      \includesvg[width=1\linewidth]{airsim_integration/airsimdemo_histories}
      \caption{Thrust Magnitude and Target Radii History}
      \label{fig:airsimdemo_results_histories}
    \end{subfigure}%

    \vspace{4mm}
    \caption{Cumulative simulation results for the most complex (right-most) test case from Fig. \ref{fig:airsimdemo_snapshots}.}

    \label{fig:airsimdemo_simresults}
\end{figure*}
    
\subsection{Experimental Setup and Simulation Environment}\label{sim_env_section}
An integrated implementation of \halss\space and \addto\space for landing a quadrotor on a Martian surface (imaged by NASA HiRISE) was created using the Microsoft AirSim plugin to Unreal Engine 4, which provides a simulated multirotor air vehicle model. Notably, the simulation vehicle is exposed to Earth's gravitational and atmospheric conditions for simplicity, however the methods proposed in this work can be extended without loss of generality to any environment and vehicle configuration (such as the Mars Helicopter \cite{balaram2018mars}). The simulation algorithms are implemented in Python (\halss) and Julia (\addto)\footnote{The code is available at \hyperlink{github.com/UW-ACL/HALO}{\url{https://github.com/UW-ACL/HALO}}}. Point clouds are generated using AirSim's body-fixed LiDAR sensor model with a maximum range of $\SI{500}{\meter}$ and a circular scanning pattern with a $30^\circ$ radial FOV.

\subsection{Integrated Simulation Results}\label{sec:integrated_sim_results}
Fig. \ref{fig:airsimdemo_snapshots} shows four different successful runs in four different environments of our coupled solution in the simulation environment\footnote{The following hyperparameters were used for all simulations: $n_{\mathrm{min}} = 3$, $n_{\mathrm{max}} = 7$, $\epsilon^j = 0.1 ~\forall j\in J$, $w_{\mathrm{des}} = [0,0,1,0,0]^\top$, $h_{\mathrm{cutoff}} = 65$ m, $\alpha = 8^\circ$, $\Delta t = 0.5$ s, $T_{\mathrm{min}} = 3.34$ N, $T_{\mathrm{max}} = 16.72$ N, $\gamma_{\mathrm{max}} = 5^\circ$, $v_{\mathrm{max}}^{\mathrm{lat}} =v_{\mathrm{max}}^{\mathrm{vert}} = 5$ m/s, and $m = 1$ kg.}. Fig. \ref{fig:airsimdemo_parametric_3D_trajectory} shows the full trajectory of the quadrotor as well as the branches of each DDTO solution in three dimensions. Notably, the initial state of the quadrotor for this simulation was above a large rock formation which caused the large divert. In Fig. \ref{fig:airsimdemo_results_histories}, to the left, we can see thrust and tilt angle control commands, which are shown to satisfy the optimization constraints (red regions). To the right, we can see the radius of each target as a function of time. Large vertical jumps in the radii correspond to previously unobserved rocks being identified as hazards within the radius of a given landing site. Smaller jumps can be attributed to hazards on the border of the circle becoming more defined and encroaching into the circle as the quadrotor descends. These results show that our coupled \halss-\addto\space solution to the HALO problem is robust to larger diverts as well as uncertainties in the landing sites.

\subsection{Comparative Study: \ddto\space vs. \addto} 
\label{subsec:comp_study_ddto_addto}
A comparative study was conducted between \ddto\space and \addto\space in a sandbox environment from a $150$ m initial altitude and targets generated in a uniform random distribution, where each target receives a $1\%$ likelihood of failure every $\Delta t$ seconds. Under 100 Monte Carlo trials, \ddto\space was observed to achieve successful landings $72.41\%$ of the time with a mean cumulative control effort of $295.05$ N-s, where as \addto\space achieved quantities of $85.56\%$ and $293.72$ N-s, respectively. This represents a $18.16\%$ increase in terms of landing success-rate and $0.4\%$ decrease in cumulative control effort from \ddto\space to \addto. 

\section{Conclusion}
\label{sec:conclusion}
In this work, we addressed the Hazard-Aware Landing Optimization  problem by introducing two novel frameworks, \halss \: and \addto, which address the challenges of perception and planning, respectively. We demonstrate the effectiveness of our solutions in a realistic simulated environment and achieve near real-time performance for both \halss \: and \addto. While this paper featured an integrated implementation, these algorithms can be used in stand-alone operations and are agnostic to the simulation environment. Moreover, our methods can apply to any application where an autonomous system has to navigate in an uncertain, dynamic, and potentially hazardous environment. 

\newpage

\bibliographystyle{IEEEtran}
\bibliography{main}  

\end{document}